%% file: egpaper_for_review.tex
\documentclass[10pt,twocolumn,letterpaper]{article}

\usepackage{cvpr}
\usepackage{times}
\usepackage{epsfig}
\usepackage{graphicx}
\usepackage{amsmath}
\usepackage{amssymb}
\usepackage{subfiles}
\usepackage{caption}
\usepackage{adjustbox}
\usepackage{booktabs}
\usepackage[ruled]{algorithm2e}
\usepackage[dvipsnames]{xcolor}
\usepackage{multido}
\usepackage{marginnote}
\usepackage{multicol}
\usepackage{multirow}

\newcommand{\stic}{STIC\xspace}

\usepackage[pagebackref=true,breaklinks=true,letterpaper=true,colorlinks,bookmarks=false]{hyperref}

 \cvprfinalcopy 

\def\cvprPaperID{11062} 

\ifcvprfinal\fi
\begin{document}

\title{Synthesize-It-Classifier: Learning a Generative Classifier through Recurrent Self-analysis}

    %

\author{Arghya Pal, Raphaël C.-W. Phan, KokSheik Wong\\
School of Information Technology, Monash University Malaysia\\
{\tt\small arghya.pal@monash.edu}
}

\twocolumn[{%
\renewcommand\twocolumn[1][]{#1}%
\maketitle
\begin{center}
    \centering
    \includegraphics[width=\textwidth,height=4.5cm]{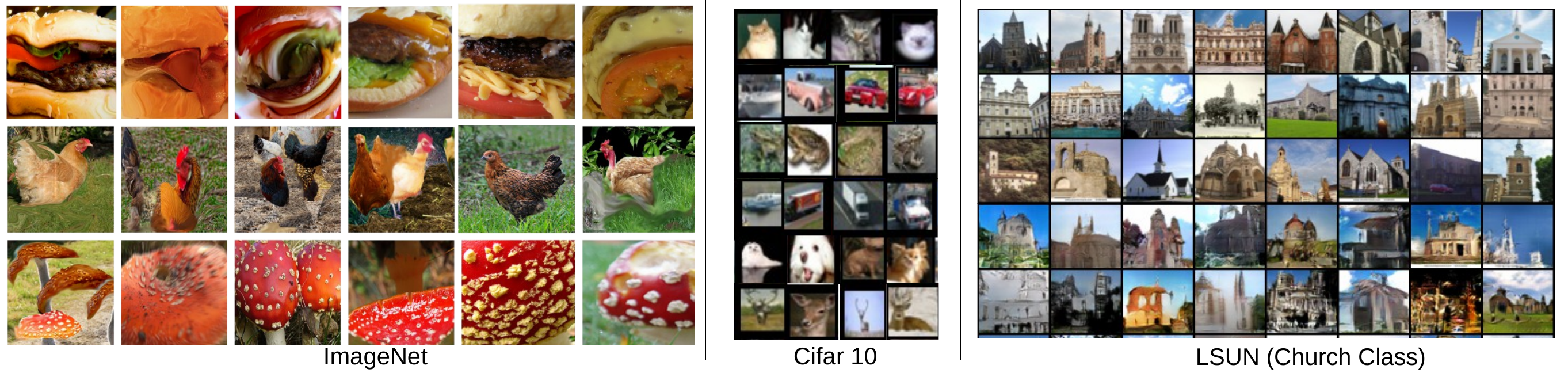}
    \captionof{figure}{\textit{\textbf{Qualitative Results of \stic Method:} (best viewed while zoomed-in):} We show qualitative results on ImageNet \cite{imagenet_cvpr09}, Cifar 10 \cite{CIFAR10} and LSUN \cite{LSUN} datasets. (1) \underline{ImageNet:} we show results of cheeseburger, chicken and mushroom classes. (2) \underline{Cifar 10:} the \stic synthesises photo-realistic images of cat, automobile, frog, truck, dog and deer classes (top-bottom rows). Variation of style (illumination, background) and content (pose, shape) can be seen for each of the classes. (3) \underline{LSUN:} We show visible geometric regularities in house shape, dome-like structures, and other outdoor entities (sky, illumination). All images are generated by using $\tau = 10$ passes. 
    The \stic methodology description is in Sec. \ref{sec_metodology}.}
     \label{fig_2_intro_images}
\end{center}%
}
]
\begin{abstract}
In this work, we show the generative capability of an image classifier network by synthesizing high-resolution, photo-realistic, and diverse images at scale. 
The overall methodology, called Synthesize-It-Classifier (STIC), does not require an explicit generator network to estimate the density of the data distribution and sample images from that, but instead uses the classifier's knowledge of the boundary to perform gradient ascent w.r.t. class logits and then synthesizes images using Gram Matrix Metropolis Adjusted Langevin Algorithm (GRMALA) by drawing on a blank canvas. During training, the classifier iteratively uses these synthesized images as fake samples and re-estimates the class boundary in a recurrent fashion to improve both the classification accuracy and quality of synthetic images. The STIC shows that mixing of the hard fake samples (i.e. those synthesized by the one hot class conditioning), and the soft fake samples (which are synthesized as a convex combination of classes, i.e. a mixup of classes \cite{zhang2018mixup}) improves class interpolation. We demonstrate an Attentive-STIC network that shows iterative drawing of synthesized images on the ImageNet dataset that has thousands of classes. In addition, we introduce the synthesis using a class conditional score classifier (Score-STIC) instead of a normal image classifier and show improved results on several real world datasets, i.e. ImageNet, LSUN and CIFAR 10. 
\end{abstract}

\begin{figure*}[t]
  \centering
  \includegraphics[height=0.25\textwidth,width=0.7\textwidth]{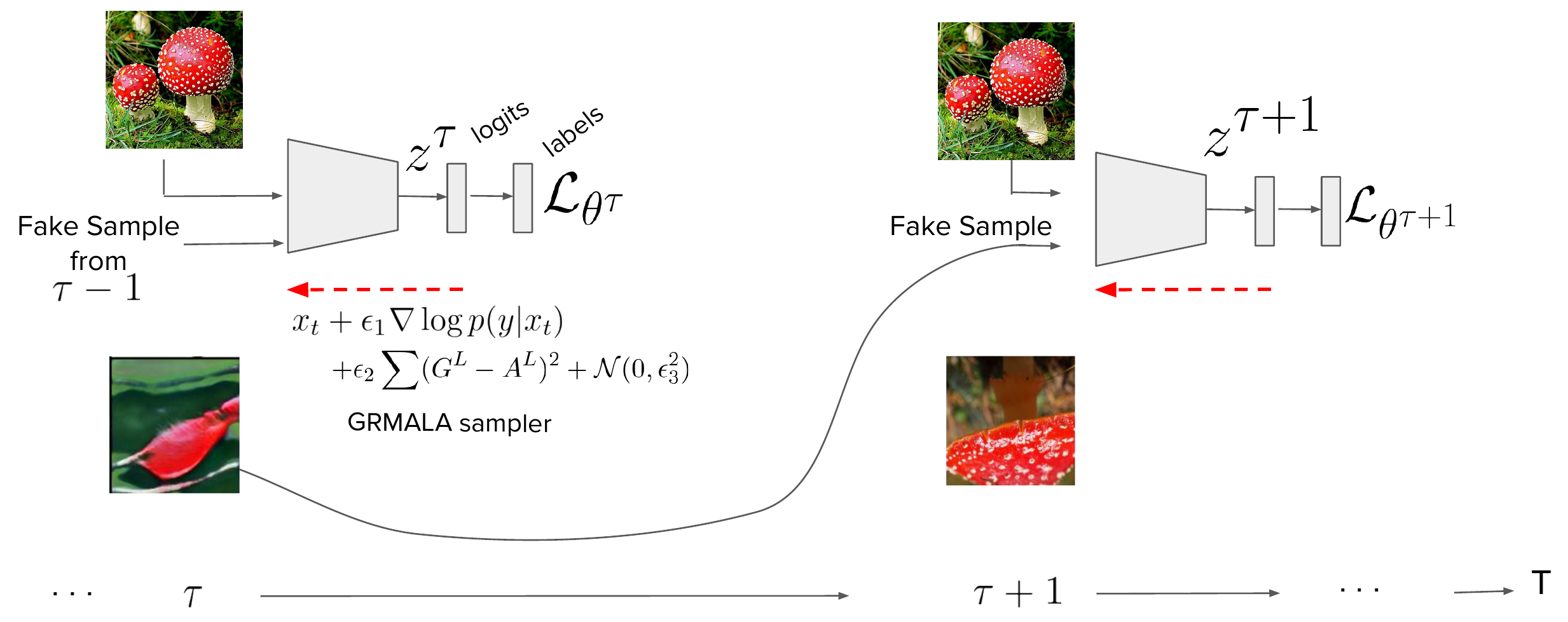}
  \caption{\textit{\textbf{The \stic Methodology:}} Our main objective is to learn a class conditional model by emphasizing the fact that $p(x|y)\propto p(y|x)$, Eqn \ref{eq_1}, and synthesise photo-realistic images from a discriminative classifier. Our proposed \stic serves dual objectives: (1) learning smooth class boundaries with Vicinal Risk Minimization; and (2) learning tighter class boundaries using recurrent self-analysis class boundary re-estimation. At time $(\tau+1)$, the classifier $p(y|x)$ is adjusting the parameters $(\theta^{\tau+1})$ using real images, mixup images; and in addition to that, synthesized images from real classes and synthesized images from mixup classes from previous iteration $\tau$ (marked as Fake Sample) are provided to the classifier. Please note that mixup classes are not actually classes but the mixup of logits of two or more classes. The samples are, at time $(\tau)$, synthesized from classifier's knowledge of the class boundary by gradient ascending w.r.t class logits, $z^{\tau}$, using our proposed Gram Matrix Regularized Metropolis Adjusted Langevin Algorithm sampler (GRMALA), see \textcolor{red}{red dashed arrow}. The \stic discriminative classifier is trained for $\tau \in \{1,2,\cdots, \mathrm{T}\}$ number of iterations.}
  \label{fig_1_Block_Diagram}
 \end{figure*}
  \vspace{-0.5cm}
\subfile{sections/1_introduction}
\subfile{sections/2_related_work}
\subfile{sections/3_methodology}
\subfile{sections/4_experiments}
\subfile{sections/5_discussion_and_analysis}
\subfile{sections/6_conclusion}
{\small
\bibliographystyle{ieee}
\bibliography{egbib}
}
\subfile{sections/7_Supl_section}

\end{document}

%% file: sections/1_introduction.tex
\section{Introduction}
\label{sec_introduction}
Discriminative classifiers $p(y|x)$ and generative models $p(x)$ are conventionally considered as domains complementary to each other, yet the distinction between them is blurring. A generative model $p(x)$ appears as a data generation process that captures the underlying density of a data distribution, whereas the discriminative classifier learns complex feature representations of images with a view to learn the class boundaries for subsequent classification. 
To elaborate, in the 
model of \cite{grathwohl2019your}, the classifier $p(y|x)$ logits are used to estimate the joint density of the image-label $p(x, y)$, and the marginal of the image distribution, $p(x)$; note the random variables, $x$: image, and $y$: class label. 
Meanwhile in \cite{jin2017introspective, lazarow2017introspective} the classifier logits are used to produce synthesized samples using an MCMC-like sampling mechanism. The classifier, on the other hand, tries to distinguish these synthesized samples and the real images to re-estimate class boundaries. We note that, synthesizing novel samples from a discriminative classifier hinges on an important factor - how well the discriminative classifier has learned the class boundaries?

We note that all the discriminative classifiers in \cite{grathwohl2019your, jin2017introspective, lazarow2017introspective, liu2019wasserstein} 
used for synthesizing novel samples are trained with Empirical Loss Minimization (ERM)~\cite{vapnik}. 
Yet, from the literature \cite{chapelle2000vicinal, zhang2018mixup}, it is evident that a discriminative classifier trained with ERM does not provide a smoother estimate of uncertainty near to the class boundary regions \cite{chapelle2000vicinal}.
Hence, we ask ourselves the question: \textit{does training with ERM have any consequence on the synthesizing capabilities of these discriminative classifiers?} We note that the transitions at class interpolation and sample quality towards class boundaries of these discriminative classifiers are neither smooth nor photo-realistic. 
In this work, we primarily seek to address this problem, \textit{viz.}, to build a discriminative classifier that will serve dual objectives: (1) the interpolated samples from one class to another must be photo-realistic; and (2) the classifier must learn tighter class boundaries so as to generate photo-realistic samples. 

To address the first objective, we train the discriminative classifier with Vicinal Risk Minimization (VRM) \cite{zhang2018mixup}. We leverage more virtual mixup
image-label samples \cite{zhang2018mixup} in addition to the real image-label samples and train the classifier. 
We then synthesize novel samples. Our novel sample synthesis method is, by design, similar 
to the Style Transfer work \cite{gatys2016image}, i.e. starting with an initial image $x_{0}$ which is updated with  gradient ascent using our proposed novel Gram Matrix Regularized Metropolis Adjusted Langavin Algorithm (GRMALA) sampler. To the best of our knowledge, this is the first discriminative classifier trained with VRM and subsequently synthesized using a novel GRMALA sampler. 
We will discuss this in detail in Sec \ref{sec_metodology}. 

Training a discriminative classifier with VRM alone, is however, a necessary 
condition for learning the smoother estimation of uncertainty among classes, but not a sufficient condition that provides tighter class boundaries. Cognitive studies \cite{anderson1991human,campbell1991if} have shown evidence where subjects (i.e. human) start with a weak cognitive decision model of an environment or the world, and recurrently refine through mistakes and self-analysis gained from the environment to develop much stronger 
cognitive decision models. In similar a spirit, we present our recurrent discriminative network trained with VRM, that we call Synthesize-it-Classifier (\stic). The \stic recurrently eliminates the regions which are outside of the class boundaries and forces the sampler to search within class boundaries. The \stic methodology trains the classifier with real images of different classes and then synthesizes samples conditioned on a class as well as the mixup samples w.r.t the class logits. At the next pass, the \stic inputs 
these synthesized samples as fake samples to the already trained discriminative classifier of the previous pass, thus allowing the classifier to re-estimate class boundaries using real images, the synthesized mixup images and the synthesized samples (we call this self-analysis). 
Similar to \cite{jin2017introspective, lazarow2017introspective}, we are, in a way, asking the classifier to quantify its own generated samples with respect to the class boundaries. The \stic does the recurrent  self-analysis for $\tau\in\{1, 2, \cdots, T\}$ number of passes.

From our empirical observations, we note that, if the image space is large (typically $> 227 \times 227$) the GRMALA sampler exhibits a slow update. We hence
show an attentive-\stic where the discriminative classifier operates on the feature space instead of raw pixel space, thus exhibiting fast update. Additionally, we also propose a novel class conditional score matching based discriminative classifier that matches the derivative of the model’s density with the
derivative of the data density \cite{song2019generative}. We will discuss each of these components elaborately in Sec \ref{sec_metodology}.

\noindent Our contributions can be summarized as follows:
\begin{itemize}
\itemsep0em 
    \item Novel recurrent self analytic \stic trained with VRM and show synthesized images using Gram matrix Regularized MALA (GRMALA) sampler w.r.t class logit
    \item We show Attentive-\stic model to address the slow mixing problem of MALA-approx. We also propose a novel class conditional score function based discriminative classifier (we call it Score-\stic method)
    \item We show results on several real world datasets, such as ImageNet, LSUN and CIFAR 10
\end{itemize}



%% file: sections/2_related_work.tex
\section{Related Work}
\label{sec_relatedwork}

\paragraph{Generative Discriminative Learning:}
The generative classifier methodology was first evident in the seminal paper ``self-supervised boosting'' \cite{welling2002self}, that learns a sequence of weak classifiers using the real data and self-generated negative samples. The use of negative samples while learning in an unsupervised manner is also seen in \cite{hinton2006unsupervised}. Similar to that, the methods in \cite{jin2017introspective, lazarow2017introspective} use the Convolutional Neural Network (CNN) based discriminative classifier’s logits and produces synthesized samples using an MCMC-like sampling  mechanism. The classifier tries to distinguish these synthesized samples and the real images to learn class boundaries. Similar to those lines of work, the method~\cite{lee2018wasserstein} shows that learning class boundaries from real and synthesized images is equivalent to optimizing the Wasserstein distance between real image and synthesized image density. Recently, the work~\cite{grathwohl2019your} shows learning of a joint distribution and a marginal distribution from the knowledge of the class logits of a discriminative classifier.
\vspace{-0.4cm}

\paragraph{Style Transfer:} There is a plethora of works that perform style transfer to meet various alternate objectives, such as: a generative adversarial learning approach to disentangle style and content of an image \cite{karras2019style}; while \cite{kotovenko2019content} propose to capture the particularity in style, and the capturing style and content of an image. The style disentanglement is shown for single image super resolution in \cite{zhang2019image}. However, in this work we will use the Gram Matrix based style transfer proposed in \cite{gatys2016image}. The seminal work of \cite{gatys2016image} computes a Gram Matrix, $G^{L} \in \mathbb{R}^{N_{L}\times N_{L}}$ using the following: the $L^{th}$ layer of a Convolutional Neural Network (CNN) has distinct $N_{L}$ feature maps each of size $M_{L} \times M_{L}$. The matrix $F^{L}\in \mathbb{R}^{N_{L}\times M_{L}}$, stores the activations $F^{L}_{i,j}$ of the $i^{th}$ filter at position $j$ of layer $L$. Then, the method computes feature correlation using: $G^{L}_{i,j} = \sum_{k} F_{i,k}^{L} F_{j,k}^{L}$, where any $F^{m}_{n,o}$ conveys the activation of the $n^{th}$ filter at position $o$ in layer $m$.
\vspace{-0.4cm}

\paragraph{Metropolis-adjusted Langevin algorithm (MALA):} The Metropolis-Hastings (MH) \cite{metropolis1953equation} uses the transition operator, \textit{viz.} $x_{t+1} = x_{t} + \mathcal{N}(0,\epsilon_{1}^{2})$, $\alpha = p(x_{t+1})/ p(x_{t})$, and if $\alpha < 1$ reject the sample $x_{t+1}$ with probability $(1-\alpha)$ and set $x_{t+1} = x_{t}$ else keep $x_{t+1}$. In practice, the MH is very slow to produce samples from any computable distribution $p_{data}(x)$. As a remedy,  \cite{roberts1998optimal, roberts1996exponential} have proposed an approximation method called Metropolis-adjusted Langevin algorithm, or the MALA. Starting with an initial $x_{0}$ typically sampled from a Gaussian distribution $\mathcal{N}(0,I)$, the MALA uses the transition operator, \textit{viz.} $x_{t+1} = x_{t} + \frac{1}{2\sigma}\nabla\log p(x_{t}) +\mathcal{N}(0,\sigma^{2})$, $\alpha = f(x_{t+1}, x_{t}, p(x_{t+1}), p(x_{t}))$, and if $\alpha < 1$ reject $x_{t+1}$ else keep $x_{t+1}$, and samples from the distribution $p(x)$. The method~\cite{nguyen2017plug} uses the stochastic gradient Langevin dynamics (SGDL) to get rid of the rejection steps of MALA and proposed the MALA-approx method. In addition to that, the method \cite{nguyen2017plug} uses different step sizes $\epsilon_{1}$ and $\epsilon_{2}$ in: $x_{t+1} = x_{t} + \epsilon_{1}\nabla\log p(x_{t}) +\mathcal{N}(0,\epsilon_{2}^{2})$ and exhibits more control over variability. In this work, we will propose a novel Gram Matrix Regularized MALA and the sampler takes the form: $x_{t+1} = x_{t} + \epsilon_{1}\nabla\log p(x_{t}) + \sum (G^{L}(x_{t}) - A^{L}(x_{t}))^{2} +\mathcal{N}(0,\epsilon_{2}^{2})$, where $\epsilon_{1}$ and $\epsilon_{2}$ are scaling factors.
\vspace{-0.4cm}

\paragraph{Vicinal Risk Minimization (VRM) using Mixup:}  
The Empirical Risk Minimization (ERM) \cite{vapnik} learns a function $f\in\mathcal{F}$ that determines the non-linear relation of the image samples $x_{i}|_{i=1}^{N}$ and the corresponding classes $y_{i}|_{i=1}^{N}$ sampled from a data distribution $p_{data}(x, y)$ by optimizing the empirical risk, $R(f) = \frac{1}{N}\sum_{i=1}^{N}l(f(x_{i}), y_{i})$. The loss function $l(\cdot)$ can be any standard loss function. Learning the function $f$ by minimizing ERM leads the function $f$ to memorize the training samples instead of a good generalization even under the purview of strong regularizer \cite{chapelle2000vicinal}. To mitigate this,  \cite{chapelle2000vicinal} proposed an alternate risk minimization technique which they refer to as Vicinal Risk Minimization (VRM), i.e. $R_{vicinity}(f) = \frac{1}{N+M}\sum_{k=1}^{N+M}l(f(\hat{x}_{k}), \hat{y}_{k})$. In VRM, we augment additional image-label pairs $(\tilde{x}_{i}, \tilde{y}_{i})|_{i=1}^{M}$ using simple geometric transformations (such as crop, rotation, mirror) of real image-label pairs $(x_{i}, y_{i})|_{i=1}^{N}$. We get the set of image-labels $(\hat{x}_{k}, \hat{y}_{k})|_{k=1}^{N+M}$ comprising augmented image-label and the real image-label pairs. The Mixup \cite{zhang2018mixup} extends this idea by augmenting virtual image-target samples, $x_{k}^{mixup} = \lambda x_{i} + (1-\lambda) x_{j}$ and $y_{k}^{mixup} = \lambda y_{i} + (1-\lambda) y_{j}$, where $\lambda\sim\text{Beta}(\alpha, \alpha)$, for $\alpha\in(0, \infty)$, also $x_{i}, x_{j}$, 
and $y_{i}, y_{j}$ are real image-labels. Mixup shows results by combining real image-label samples of different classes instead of hand-crafted data augmentation of images. The VRM of Mixup can be defined as, $R_{mixup}(f) = \frac{1}{N+K}\sum_{l=1}^{N+K}l(f(x_{l}), y_{l})$. We get the set of image-labels $(x_{l}, y_{l})|_{l=1}^{N+K}$ from the real image-label pairs and mixup image-label pairs. In this work, we will use the image-labels $(x_{l}, y_{l})|_{l=1}^{N+K}$ set.

%% file: sections/3_methodology.tex
\section{The \stic Methodology}
\label{sec_metodology}
In this work, we wish to learn the parameter of a class conditional distribution of image $x$ and the corresponding class label $y$ (we fix $y$ to be from a particular class $y_{c}$), i.e.: 
\begin{equation}
\label{eq_1}
    p(x|y=y_{c})
\end{equation}
with a view to generating photo-realistic novel samples. 

We can expand the class conditional model in Eq \ref{eq_1} using the Bayes rule, i.e.: $p(x|y) = p(x) p(y|x)/p(y)\propto p(x) p(y|x)$. We, however, cannot directly write a sampler by utilizing the ``product of experts'' \cite{hinton1999products}, as we do not have a generator network $p(x)$ in our setup. Since the random variable $y$ is categorical, we instead can write a modified version, i.e:
\vspace{-0.4cm}

\begin{equation}
\label{eq_stic_original}
 \begin{split}
    & p(x|y) = p(x) p(y|x)/p(y) \propto p(y|x)
\end{split}
\end{equation}
\noindent such that, estimating the density directly has a relation to how well synthesized samples are classified by the discriminative classifier network. 

Following the Style Transfer work \cite{gatys2016image} and the sampling with Langevin algorithm work in \cite{nguyen2017plug, roberts1998optimal}, we propose a Gram Matrix Regularized MALA approx (GRMALA) sampler and propose the following update rule for $x_{t+1}$:
\vspace{-0.4cm}

\begin{equation}
\label{eq_grmala}
\begin{split}
x_{t} + \epsilon_{1}\nabla\log p(y|x_{t}) + \epsilon_{2}\sum (G^{L} - A^{L})^{2} +\mathcal{N}(0,\epsilon_{3}^{2})
\end{split}
\vspace{-0.5cm}
\end{equation}
\noindent and, similar to MALA-approx proposed in \cite{nguyen2017plug}, we use different step sizes, i.e. $\epsilon_{1}, \epsilon_{2}, \epsilon_{3}$ for three terms after $x_{t}$ in Eq \ref{eq_grmala}. Here, $\epsilon_{1}$ and $\epsilon_{2}$ controls the sample quality and $\epsilon_{3}$ controls the diversity by moving around the search space. Note that we get the Gram Matrix $G^{L}$ from the $x_{t}$ and we get Gram Marix $A^{L}$ from a real image $x$ (for more details on Gram Matrix please refer to \cite{gatys2016image}, or Sec \ref{sec_relatedwork} Style Transfer section). In order to generate photo-realistic synthsized images, our discriminative classifier, hence, must serve two objectives: (1) \textit{Learning Smooth Class Boundaries using VRM} such that the interpolated samples from one class to another must be photo-realistic; and (2) \textit{Learning of Tighter Class Boundaries using Recurrent Self-analysis Class Boundary Re-estimation} such that the classifier must learn tighter class boundaries so as to generate photo-realistic samples. 
\vspace{-0.4cm}

\paragraph{Learning Smooth Class Boundaries using VRM:} Similar to \cite{zhang2018mixup}, we augment mixup image-label pairs along with real image-label pairs. We have $K$ number of mixup augmented image-label pairs $(x^{mixup}_{k}, y^{mixup}_{k})|_{k=1}^{K}$, those we get after, $x_{k}^{mixup} = \lambda x_{i} + (1-\lambda) x_{j}$ and $y_{k}^{mixup} = \lambda y_{i} + (1-\lambda) y_{j}$, where $\lambda\sim\text{Beta}(\alpha, \alpha)$, for $\alpha\in(0, \infty)$, also $x_{i}, x_{j}$, and $y_{i}, y_{j}$ are real image-label pairs. For brevity, let us assume that the mixup image-label pairs are coming from a mixup distribution $(x^{mixup}_{k}, y^{mixup}_{k})\sim p_{mixup}(x^{mixup}, y^{mixup})$ and we have our real image-label distribution $(x_{i}, y_{i}) \sim p_{data}(x,y)$. Our objective function to optimize Eq \ref{eq_stic_original} is the following:
\begin{equation}
\label{eq_vrm}
\begin{split}
    & \mathcal{L}(\theta) = - \sum_{(x_i, y_i)\sim p_{data}}^{i=1,\cdots,N} \log p_{\theta}(y_i = y_{c}|x_i) \\
    & \qquad - \sum_{(x^{mixup}_{k}, y^{mixup}_{k})\sim p_{mixup}}^{k=1,\cdots,K} \log p_{\theta}(y_{k}=y_{mixup}|x^{mixup}_{k})
\end{split}
\end{equation}
\noindent 
where we note here that $y_{k} = y^{mixup}$ is not a true class but represents the mixing of true class logits.  
\vspace{-0.4cm}

\paragraph{Learning of Tighter Class Boundaries using Recurrent Self-analysis Class Boundary Re-estimation:} Learning smooth class boundaries using VRM is a necessary condition for smooth image synthesis but not a sufficient condition for learning tighter class boundaries with a view to synthesize photo-realistic images. We hence introduce a recurrent self-analysis class boundary re-estimation methodology that eliminates the regions which are outside of the class boundaries and force the sampler to focus 
within the class boundaries. To achieve this objective, we now describe a recurrent training procedure that spans around $\tau \in \{1,2,\cdots, \mathrm{T}\}$ number of passes. At pass $\tau$, we synthesize novel samples from a trained classifier $p_{\tau}(\cdot)$ by GRMALA based update with respect to the class logits. At the next pass, $\tau + 1$, the \stic takes images from dataset and mixup images as real images. On the other hand, synthesized images of real classes and synthesized images of mixup classes from the classifier at pass $\tau$ are taken as fake samples (note that such synthesized samples are taken from the trained classifier at previous pass $\tau$, see Figure \ref{fig_1_Block_Diagram} fake images). Thus allowing the classifier to re-estimate class boundaries using the real images, the synthesized mixup images and the synthesized samples. 
We call this a recurrent self-analysis. 
The recurrent class boundary re-estimation is, in a way, asking the classifier to quantify its own generated samples with respect to the class boundaries. 
We sample and re-train the classifier for $\tau \in \{1,2,\cdots,\mathrm{T}\}$ times, thus enabling the classifier to re-estimate its class boundaries at each time step. 
For the $(\tau+1)^{th}$ time step, the objective function of the classifier hence then becomes:
\begin{equation}
\begin{split}
    & \mathcal{L}(\theta^{\tau+1}) = - \sum_{(x_i, y_i)\sim p_{data}}^{i=1,\cdots,N} \log p_{\theta^{\tau+1}}(y_i = y_{c}|x_i)\\
    & \quad - \sum_{(x^{mixup}_{k}, y^{mixup}_{k})\sim p_{mixup}}^{k=1,\cdots,K} \log p_{\theta^{\tau+1}}(y_{k}=y_{mixup}|x^{mixup}_{k})\\
    & \quad - \sum_{(x_i, y_i)\sim p_{\theta^\tau}}^{i=1,\cdots,N} \log p_{\theta^{\tau+1}}(y_i = -1 |x_i)\\
    & \quad - \sum_{(x^{mixup}_{k}, y^{mixup}_{k})\sim p_{\theta^\tau}}^{k=1,\cdots,K} \log p_{\theta^{\tau+1}}(y_{k}= -1|x^{mixup}_{k})
\end{split}
\end{equation}

Theoretically, the softmax of the classifier $p_{\theta}^{\tau+1}(y|x)$ is: $\frac{\exp(p_{\theta}^{\tau + 1}(x)[y])}{\sum_{y^{'}}\exp(p_{\theta}^{\tau+1}(x)[y^{'}])}$. 
Hence, we can approximate the $p(x,y)$ by following, $p_{\theta}^{\tau+1}(x,y) = \exp(p_{\theta}^{\tau}(x)[y])/Z(\theta)$. 
Please note that, we get $p_{\theta}^{\tau}(\cdot)$ from previous time step $\tau$. 
Marginalizing $y$ from $p_{\theta}^{\tau+1}(x,y)$, i.e. $p_{\theta}^{\tau+1}(x)=\sum_{y} p_{\theta}^{\tau+1}(x,y)= \sum_{y} \exp(p_{\theta}^{\tau}(x)[y])/Z(\theta)$ provides us the estimation of $p(x)$. 
However, $p(x)$ is dropped from Eqn \ref{eq_stic_original} as there is no explicit network and the learning is incorporated through GRMALA and $p_{\theta}^{\tau}(\cdot)$.

%% file: sections/4_experiments.tex
\begin{figure*}[t]
  \centering
  \includegraphics[height=0.2\textwidth,width=0.9\textwidth]{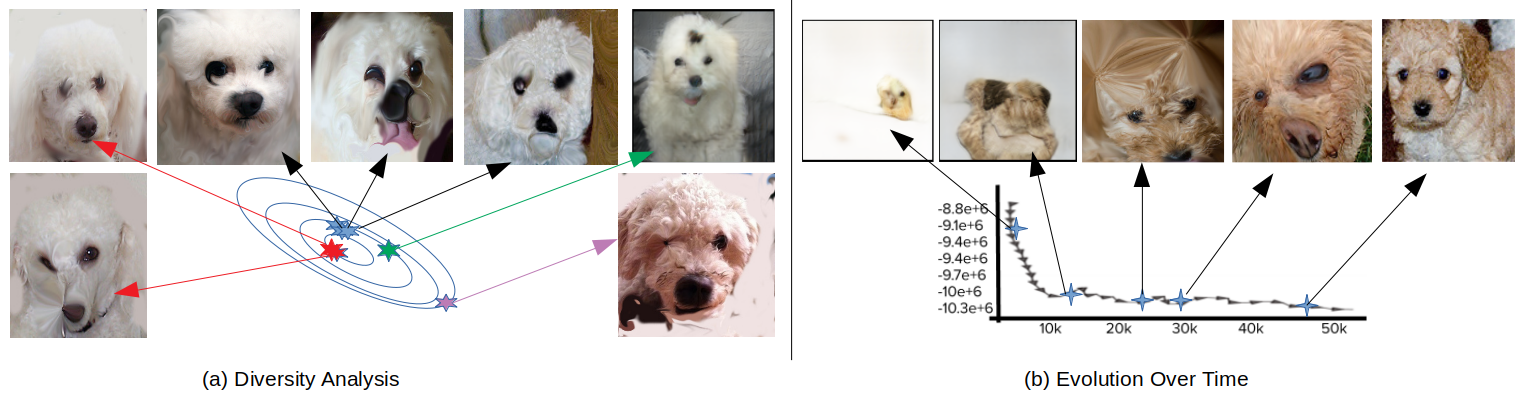}
  \caption{\textit{\textbf{(a) Diversity Analysis:}} we synthesize samples from one class and samples from the neighborhood around those samples to get other starting samples on ImageNet class dog. We note, samples which are in near vicinity show similar object appearance (observe same face structures of black arrow samples). In contrast, samples which are far apart (see red arrow and purple arrow samples) show different appearance of the same dog class. \textit{\textbf{(b) Evolution Over Time:}} we show class dog synthesized samples of ImageNet at different iterations, i.e. $\{10k, 20k, \cdots, 50k\}$ (horizontal axis: no. of iterations, vertical axis: training loss). Images are blurry initially but become clearer over time, showing that the proposed method is learning tighter class boundaries over the time steps.}
  \label{fig_3_diversity_evolution}
  \vspace{-0.45cm}
  
\end{figure*}
\section{Experiments and Results}
\label{sec_experiments_and_results}
We perform a comprehensive suite of experiments and ablation studies, across standard benchmark datasets; specifically on three standard datasets: ImageNet \cite{imagenet_cvpr09}, Cifar 10 \cite{CIFAR10} and LSUN \cite{LSUN}.
\vspace{-0.45cm}

\paragraph{Baseline and SOTA methods:} By design, our method is a hybrid network that can simultaneously perform classification and synthesis. From the class conditional generative network end, we observe that the BigGAN \cite{brock2018large}, PnP \cite{ nguyen2017plug}, SNGAN \cite{miyato2018spectral} methods are state-of-the-art (SOTA) for class conditional image generation. In terms of the generative discriminative learning, the works of JEM \cite{grathwohl2019your}, INN \cite{jin2017introspective}, WINN \cite{ liu2019wasserstein}, EBM \cite{xie2016theory} are closer to our work. However, our proposed \stic, to a large extent, differs from these methods as follows: (1) the crucial difference is that our discriminative classifier is trained with VRM, and (2) we use a novel Gram Matrix MALA sampler. 
We consider BigGAN-deep (res 256, channel 96, parms 158.3, shared, orthogonal reg, skip-z)~\cite{brock2018large}, cascade classifier network model from \cite{jin2017introspective, liu2019wasserstein} methods, and other methods as described in their corresponding paper. 
While for classifiers, we consider ResNet \cite{he2016deep}, MobileNet \cite{howard2017mobilenets}, and GoogleLenet (GLent) \cite{szegedy2015going} as the SOTA methods then compare our method against these SOTAs. 
We consider INN~\cite{lazarow2017introspective} as our baseline method for synthesizing method, as we note that such earlier effort uses a discriminative classifier to synthesize novel samples from its understanding of class boundary information. 
These synthesized samples and real images are then utilized by INN method for class boundary re-estimation. 
For discriminative classifier, we use GoogleLeNet as our baseline method. 
Here, a batch size of 50 is considered for all SOTA methods unless specified otherwise.
\vspace{-0.45cm}

\paragraph{Network Setup and Hyperparameter Choices of \stic:} Similar to the previous work~\cite{grathwohl2019your}, 
we use a Wide Residual Network \cite{zagoruyko2016wide}, WideResNet-28-10, without batch-normalization to make \stic output deterministic functions of the input. 
The Adam optimizer, $5k$ iteration for each pass $\tau \in \{1,2,\cdots, 10\}$ totaling $50k$ iterations, the Langevin dynamics chains are evolved after 15 epochs (after one pass) and with probability 0.5 we re-initialize the chains with uniform random noise. For pre-processing, we scale images to the range $[-1, 1]$ and add Gaussian noise of stddev = 0.3 (owing to the page constraint more description is deferred to Supplementary section). We have two notions for time, a pass $\tau$ and iteration: we start training, at pass $\tau = 1$. At pass $\tau=1$, the classifier with real images, virtual mixup images, while considering blank images (pixel intensities are set to 255) as fake images. One pass continues for $5k$ iterations and then we synthesize fake images from the classifier $p_{\theta^{1}}(y|x)$. We then move to the next pass $\tau=2$ that lasts for another $5k$ iterations. We have a total number of 10 passes, i.e. $50k$ iterations, for \stic training.
\vspace{-0.45cm}

\paragraph{Qualitative Results:}
Sample labeled image generations of the proposed \stic method are summarized in Fig \ref{fig_2_intro_images}. The zoomed-in versions of those images and more qualitative results are presented in the Supplementary Section. Note that \stic generates images with improved quality in multiple cases across the datasets. In LSUN, proper geometric shapes for house and sky of synthesized images by \stic; in ImageNet and in Cifar 10 synthesized images, we observe style and content information are captured by \stic.
\begin{figure*}[t]
  \centering
  \includegraphics[height=0.2\textwidth,width=0.87\textwidth]{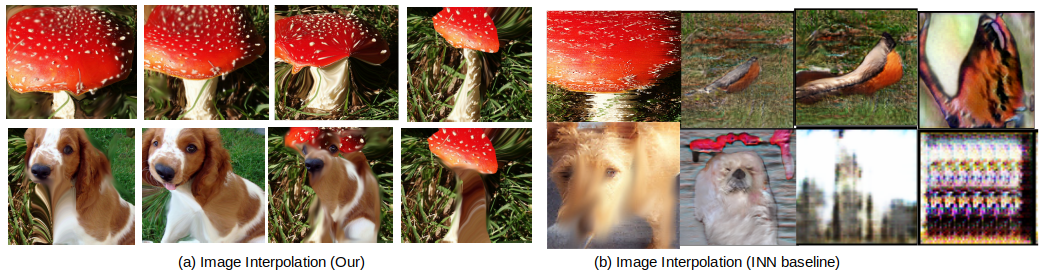}
  \caption{\textit{(a) \textbf{Image Interpolation} (ours)} first four columns show image interpolation result of our method. We notice smooth transition from one class $c_1$ to another class $c_2$. \textit{(b) \textbf{Interpolation of Result of INN} (baseline):} We note that the class interpolation from one class to other is not smooth, i.e. in-between images are not human interpretable.}
  \label{fig_4_interpolation_loss}
\end{figure*}
\vspace{3pt}

\begin{figure}[!h]
  \centering
  \includegraphics[height=0.17\textwidth,width=0.4\textwidth]{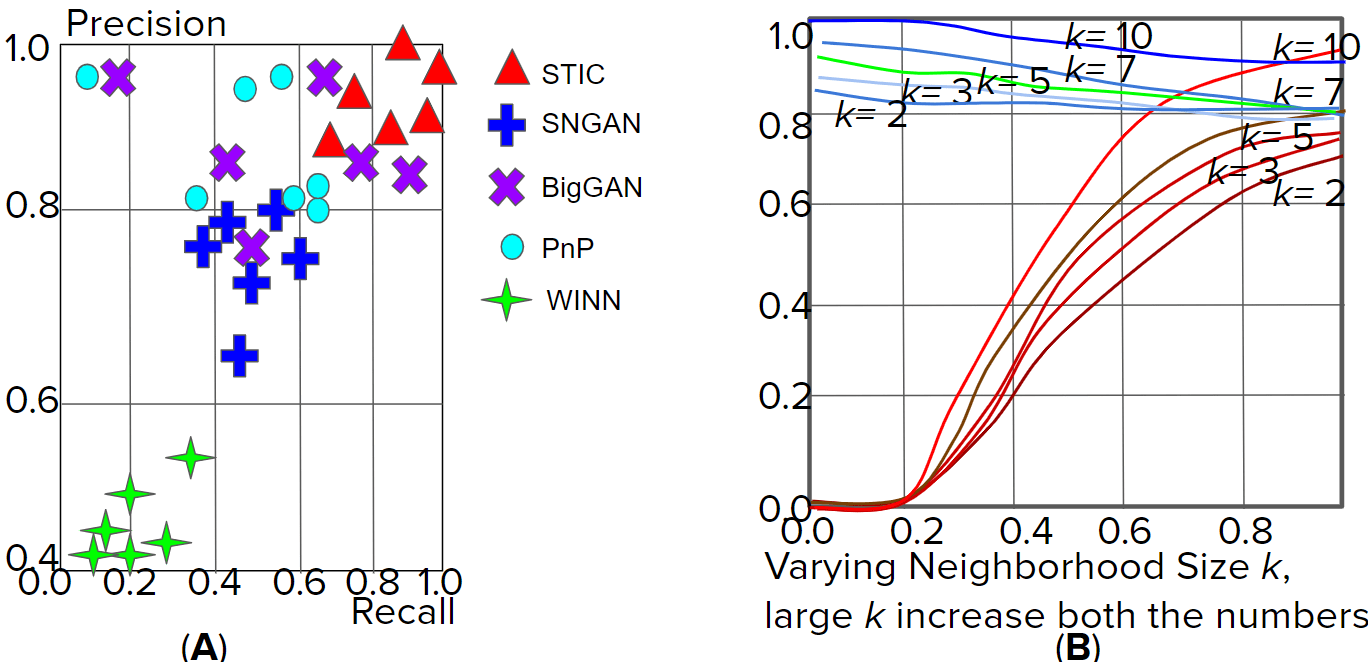}
  \caption{\textbf{Generalizability of \stic Method:} \textbf{(A)} We show the precision-recall comparison of \stic, SNGAN, BigGAN, PnP and WINN at different initializations. A high precision-recall for \stic justifying our claim. \textbf{(B)} Precision-recall at different $k$-NN using the features of a ResNet50 classifier.}
  \label{fig_P_R_K}
\end{figure}

\noindent \textit{Diversity Analysis:} At pass $\tau$, we synthesize class conditioned sample $p_{\theta^{\tau}}(x|y=y_{c_{1}})$ of class $y_{c_{1}}$ (see black arrows in Fig \ref{fig_3_diversity_evolution} (a)). Similar to marginal density estimation proposed in \cite{grathwohl2019your}, we use a small neighborhood around $p_{\theta^{\tau}}(x|y=y_{c_{1}})$ as other starting samples to understand the capability of the model to generate diverse samples. It is evident that samples which are in near vicinity show similar object appearance (observe same face structures of black arrow samples in Fig \ref{fig_3_diversity_evolution} (a)), similar background (observe similar facial structure and background in red arrow samples in \ref{fig_3_diversity_evolution} (a)). In contrast, samples which are far apart, for example, see red arrow and purple arrow samples in \ref{fig_3_diversity_evolution} (a), show different appearance of the same dog class.
\vspace{3pt}

\noindent \textit{Latent Space Interpolation:} Two points $p(x|y=y_{c_1})$ and $p(x|y=y_{c_2})$ are sampled from two distinct classes $c_1$ and $c_2$ at pass $\tau=10$ and then linearly interpolate between $p(x|y=y_{c_1})$ and $p(x|y=y_{c_2})$ to obtain novel samples. 
The synthesized images of ImageNet are shown in Fig \ref{fig_4_interpolation_loss}(a). 
Synthesized images from one class to another are smooth and human interpretable, as opposed to the interpolation provided by the baseline INN \cite{jin2017introspective} in Fig \ref{fig_4_interpolation_loss} (b), i.e. in-between images are not human interpretable. 
Thus supporting our claim that \stic provides smooth synthesised samples.
\vspace{3pt}

\noindent \textit{Evolution over Time Steps:} 
In Fig \ref{fig_3_diversity_evolution} (b), we show the qualitative results of class dog of ImageNet at different iterations, i.e. $\{10k, 20k, \cdots, 50k\}$. Please note that, in \stic setup, $5k$ iteration stands for one pass of $\tau \in \{1,2, \cdots, 10\}$. The generated images are blurry initially but become clearer over time, showing that the proposed method is learning tighter class boundaries over the time steps.
\vspace{3pt}

%
\begin{table*}[!ht]
\centering
\begin{adjustbox}{width=.7\textwidth,center}
\begin{tabular}{lllllllllllll}
\hline
\multirow{1}{*}{Methods} & \multicolumn{4}{c}{LSUN} & \multicolumn{4}{c}{CIFAR10} & \multicolumn{4}{c}{ImageNet} \\ \cline{2-13} 
& \textbf{MIS} & \textbf{FID} & \textbf{Cls}$_{R}$ & \textbf{Cls}$_{G}$ & \textbf{MIS} & \textbf{FID} & \textbf{Cls}$_{R}$ & \textbf{Cls}$_{G}$ & \textbf{MIS} & \textbf{FID} & \textbf{Cls}$_{R}$ & \textbf{Cls}$_{G}$\\ 
 & $(\uparrow)$ & $(\downarrow)$ & $(\uparrow)$ & $(\uparrow)$ & $(\uparrow)$ & $(\downarrow)$ & $(\uparrow)$ & $(\uparrow)$ & $(\uparrow)$ & $(\downarrow)$ & $(\uparrow)$ & $(\uparrow)$ \\\hline
INN & 14.91 & 45.62 & 26 & 10 & 0.93 & 118.92 & 29 & 20 & 1.92 & 189.05 & 52 & 30 \\ 
WINN & 17.43 & 38.03 & 41 & 28 & 21.94 & 51.81 & 48 & 36 & 21.13 & 58.72 & 48 & 38 \\
PnP & 32.03 & 15.07 & 62 & 58 & 31.37 & 17.93 & 54 & 53 & 33.18 & 14.71 & 61 & 54 \\ 
JEM & 28.92 & 40.42 & 60 & 39 & 38.4 & 47.60 & 57 & 39 & 32.32 & 40.41 & 53 & 32 \\ 
EBM & 31.83 & 19.73 & 62 & 50 & 31.63 & 17.02 & 58 & 50 & 32.81 & 30.90 & 63 & 52 \\ 
BigGAN & \textbf{113.13} & \textbf{8.67} & 88 & 87 & 100.31 & \textbf{7.92} & 89 & 81 & 99.31 & \textbf{8.51} & 85 & 80 \\ 
SNGAN & 52.37 & 17.43 & 61 & 59 & 53.01 & 20.3 & 83 & 78 & 65.72 & 12.62 & 67 & 61 \\\hline
\underline{\stic} & \underline{93.61} & \underline{13.32} & \underline{96} & \underline{92} & \underline{97.91} & \underline{12.81} & \underline{91} & \underline{90} & \underline{98.62} & \underline{15.01} & \underline{95} & \underline{93} \\ 
\underline{\stic-ERM} & \underline{30} & \underline{35.92} & \underline{72} & \underline{62} & \underline{20} & \underline{48.17} & \underline{61} & \underline{60} & \underline{27.19} & \underline{38.27} & \underline{65} & \underline{63} \\ 
\underline{Attentive-\stic} & \underline{99.61} & \underline{9.01} & \underline{97} & \underline{95} & \underline{100.56} & \underline{11.71} & \underline{93} & \underline{90} & \underline{100.19} & \underline{10.38} & \underline{96} & \underline{93} \\ 
\underline{Score-\stic} & \underline{112.61} & \underline{8.82} & \underline{\textbf{98}} & \underline{\textbf{96}}& \underline{\textbf{108.62}} & \underline{9.99} & \underline{\textbf{97}} & \underline{\textbf{92}} & \underline{\textbf{104.91}} & \underline{8.83} & \underline{\textbf{97}} & \underline{\textbf{95}} \\ \hline
ResNet & N/A & N/A & 80 & 73 & N/A & N/A & 69 & 67 & N/A & N/A & 67 & 63 \\ 
WideResNet & N/A & N/A & 83 & 67 & N/A & N/A & 79 & 71 & N/A & N/A & 67 & 63 \\ 
MobileNet & N/A & N/A & 87 & 77 & N/A & N/A & 89 & 68 & N/A & N/A & 87 & 83 \\ 
GLent & N/A & N/A & 88 & 71 & N/A & N/A & 83 & 77 & N/A & N/A & 86 & 80 \\ \hline
\end{tabular}%
\end{adjustbox}
\caption{\textbf{Quantitative Results of Various Real-world Image Datasets:} We report: (i) \textbf{MIS} ($\uparrow$, higher is better); (ii) \textbf{FID} ($\downarrow$, lower is better); (iii) \textbf{Cls}$_{R}(\uparrow$, higher is better); and (iv) \textbf{Cls}$_{G}(\uparrow$, higher is better). We mark winning entries in bold. The \stic and its variants are underlined. The N/A stands for not applicable.}
\label{table_quantitative_results}
\vspace{-0.45cm}

\end{table*}

\noindent \textit{Quantitative Evaluation:}
We used multiple quantitative metrics to study the proposed method on generated image quality, diversity and image-label correspondence: (i) \textbf{MIS} ($\uparrow$, higher is better) \cite{gurumurthy2017deligan}; (ii) \textbf{FID} ($\downarrow$, lower is better) \cite{heusel2017gans}; (iii) \textbf{Cls}$_{R}(\uparrow$, higher is better), i.e. Top-5 classification accuracy (in $\%$) of a ResNet-50 classifier trained on real labeled images and tested on generated images; and (iv) \textbf{Cls}$_{G}(\uparrow$, higher is better), i.e. Top-5 classification accuracy (in $\%$) of a ResNet-50 classifier trained on generated/synthesized labeled images and tested on real images. The results are shown in Table \ref{table_quantitative_results}. We observe a distinct performance gain for \stic over the state-of-the-art models. The low FID score and high \textbf{Cls}$_{G}$-based classification accuracy scores imply diverse image-label generation. In particular, the improved classification performance, as shown through \textbf{Cls}$_{T}$ and \textbf{Cls}$_{G}$, demonstrate the utility of the synthesized labeled images for downstream classification tasks.

\noindent \textit{Classification Accuracy improvement with \stic:} In Table \ref{table_quantitative_results}, we show that \stic improves not only the generation quality but also the classification accuracy of the Wide ResNet classifier. 
It it worthy that the \stic classifier not only improves the Wide ResNet classifier but it also achieves the highest \textbf{Cls}$_{R}$ and \textbf{Cls}$_{G}$ scores with respect to GoogleLenet (GLent in Table \ref{table_quantitative_results}) and MobileNet. 
This shows that the recurrent self-analysis obtains tighter class boundaries.

%% file: sections/5_discussion_and_analysis.tex
\begin{figure*}[t]
    \centering
    \includegraphics[width=\textwidth, height=0.3\textwidth]{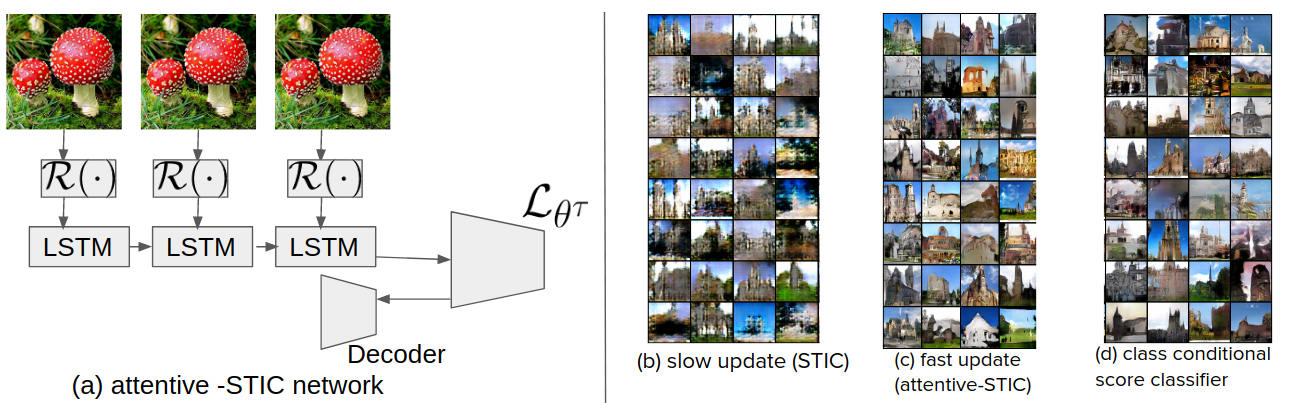}
    \caption{\textit{(a-c) \textbf{attentive-\stic}:} \stic method can work in feature space. We show qualitative results of \stic and attentive-\stic on LSUN church at $10k$ iteration and note improved results. \textit{(d) \textbf{score-\stic}:} we show the qualitative results of score-\stic only after $10k$ iterations. We show geometric details on these LSUN church samples.}
    \label{fig:5_attentive-STIC}
\end{figure*}
%
\vspace{-0.2cm}

\section{Discussion and Analysis}
\label{sec_discussion}
\vspace{-0.2cm}

\paragraph{Discussion of Quantitative Results:}
From Table \ref{table_quantitative_results}, we note that INN, WINN do not perform well due to training with ERM and learning from a weaker classifier. The PnP performance drops due to the apparent complexity while training the prior network. The \stic methodology supports the primary claim of a deep generative model of benefiting downstream tasks, such as classification. We, hence, see that the classifier in \stic methodology learns a tighter decision boundary (see improved \textbf{Cls}$_{R}$ and \textbf{Cls}$_{G}$) and smooth class interpolation to achieve this objective. However, the FID calculates the distance between feature vectors of real and generated images. We note that the classifier in \stic methodology learns a tighter decision boundary may not learn a good feature similarity of real and fake images, hence a slight drop in FID score w.r.t BigGAN. For classifier networks, we note a performance boost w.r.t SOTA classifier networks, thus showing the efficacy of our methodology as a classifier.
\vspace{-0.4cm}

\paragraph{Generalizability of \stic Model:} To understand the generalizability of the \stic method we adopt the precision-recall and $k$-nearest neighbor (KNN) analysis proposed in \cite{NEURIPS2019_0234c510}. Fig \ref{fig_P_R_K} (a) shows high precision and recall at different initilizations, thus supporting our claim of diversity and generalizability in Sec \ref{sec_experiments_and_results}. Similarly, we show precision and recall at different KNN using features of ResNet-50.
\vspace{-0.4cm}

\paragraph{Ablation of Gram Matrices}: In this work, we use the style representation of deeper layers, `conv21'-`conv28' of \stic model and got FID: 15.01 on ImageNet. To show the effectiveness of style transfer from the deeper layers we do style transfer from shallow layers `conv1'-`conv20' and that gives FID: 28 on ImageNe, thus not capturing more style. Howvever, considering all layers `conv1'-`conv28' FID:30, mixes learning of deeper style and shallow layer style, thus leading to bad FID.
\vspace{-0.4cm}

\paragraph{Running Time Complexity:} Training our model was $\sim 3.2 \times$ faster than training BigGAN and SNGAN. This is primarily because of the time taken for stabilization of GANs during training. Similarly, \cite{nguyen2017plug} optimizes two separate networks making their training time significantly larger. Also, INN \cite{jin2017introspective} and WINN \cite{lee2018wasserstein} trains multiple classifiers in a sequence ($>25$ number of classifiers in a sequence) for a single image synthesis, making its overall synthesis costly.
\begin{figure}
    \centering
    \includegraphics[height=0.3\textwidth]{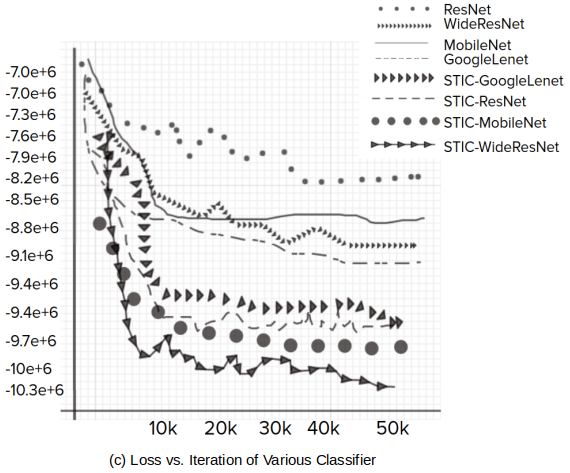}
    \caption{We show Loss (vertical axis) \textit{vs.} No. of Iterations (horizontal axis) of discriminative classifier methods. We note that, classifiers performances improved after adopting \stic methodology}
    \label{fig_loss}
    \vspace{-0.8cm}
\end{figure}
\vspace{-0.4cm}

\paragraph{Effect of \stic on Other SOTA Classifiers:} We ask ourselves whether a recurrent self-analysis method improves the classification accuracy of any classifier? We answer this in Fig \ref{fig_loss}. We show loss per iteration and show that the \stic methodology improves the training accuracy of any classifier.
\vspace{-0.4cm}

\paragraph{Optimal Number of Passes:} In Sec \ref{sec_experiments_and_results} we show results for $\tau=10$ number of passes. In thise section, we will study the number of passes and their relation with FID and other scores. We found that beyond $\tau=10$ number of passes the synthesized image quality the FID and MIS scores minimally improves. Improving FID and MIS scores by leveraging more advanced classifers could be a possible future direction.
\vspace{-0.4cm}

\paragraph{Attentive-\stic to Mitigate Slow Update of GRMALA:} The MALA-approx sampler often results in a slow mixing while dealing with high dimensional pixel space. The major drawback of slow mixing is the modeling of the data distribution. To ameliorate the problem, we will use GRMALA in the feature space instead of the pixel space. Motivated by the improved performances of attention based encoding \cite{gregor2015draw}, we use an attention based feature encoder comprised of: (1) a reading network, $\mathcal{R}(\cdot)$ that receives an image $\textbf{x}$ and decides to focus on a part of $\textbf{x}$ using an attention mechanism (described later); (2) the $\mathcal{R}(\cdot)$ then outputs a vector $\textbf{v}_{t}$ (which is rasterized from the patch being attended to); (3) an LSTM network receives $\textbf{v}_{t}$ and provides a feature vector $f$. Similar to the DRAW \cite{gregor2015draw} reading mechanism:$\hat{x}_{t} = x -  \zeta(\hat{x}_{t-1}), v_{t} = \mathcal{R}(x,\hat{x}_{t}, v_{t-1}); [f_{t}, h^{enc}_{t}] = LSTM(v_{t}, h^{enc}_{t-1})$, here, $\zeta(\cdot)$ is a sigmoid function. The classifier, $p(y=y_{c}|f)$, now operates on the extracted feature of an image $x$ and synthesize feature vector. The synthesize vector is passed to decoder network (see Fig \ref{fig:5_attentive-STIC}) to upsample the feature vector to get synthesized image. The decoder is the DCGAN network. We show the qualitative results on LSUN church classes after one pass $\tau=1$ (i.e. $5k$ iterations), please see network in Fig \ref{fig:5_attentive-STIC} (b) for results. In addition to that, the quantitative results are shown in Table \ref{table_quantitative_results}.
\vspace{-0.4cm}

\paragraph{Score-\stic a Class Conditional Score Discriminative Classifier:} based on our understanding from Eqn \ref{eq_stic_original}, the \stic method depends on discriminative classifier. To this end, we propose a small modification on Wide ResNet architecture (or, modification to any classifier network in general). The \cite{song2019generative} method attempts to match the derivative of the model’s marginal density with the derivative of the marginal density of real data using a score of a probability density $p(x)$, i.e. $\nabla_{x}\log p(x)$. We extended this idea and propose a novel class conditional score based Wide ResNet that we refer score-\stic. The WideResNet-28-10 last layer dimension is matched with input layer dimension (which is a criteria for score network \cite{song2019generative}) followed by softmax classification. The following equation acts as a regularizer to the Eqn. \ref{eq_stic_original}, i.e.: $\mathbb{E}_{p_{data}(x)}\big[\frac{1}{2}||p_{\theta^{\tau}}(x)||_{2}^{2} + tr(\nabla_{x}p_{\theta^{\tau}}(x)) + \frac{1}{2}||(y_{c}, p_{\theta^{\tau}}(y|x))||_{2}^{2}\big]$. We show results in Fig \ref{fig:5_attentive-STIC} and Table \ref{table_quantitative_results}.

%% file: sections/6_conclusion.tex
\vspace{-0.3cm}

\section{Conclusion}
\label{sec_conclusion}
\vspace{-0.2cm}

In this work, we emphasize on the relation $p(x|y) \propto p(y|x)$ and propose \stic method to synthesize images using Gram-matrix Regularized MALA (GRMALA) sampler w.r.t class logit. Our classifier satisfies: (1) smooth interpolation; and (2) a tighter class boundaries so as to generate photo-realistic samples. To this end, we propose a novel recurrent self-analytic \stic trained with VRM. We further show an Attentive-\stic model to address the slow mixing problem of GRMALA. In addition to that, we show a novel class conditional score function based Wide ResNet classifier and show improved generation. We present results on several real world datasets, such as ImageNet, LSUN and Cifar10.

%% file: sections/7_Supl_section.tex
\clearpage

\begin{center}
\large{\textbf{Supplementary Section}}
\end{center}
In this section, we include more details which could not be included in the main paper due to space constraints.
\begin{itemize}
\itemsep0em
    \item \textbf{Architecture Details and Algorithm:} discussion, analysis on the architecture, and pseudo code of the \stic methodology are provided in Sec \ref{suppl_architecture_details}.
    \item \textbf{\stic vs. SOTA Image Synthesis using Discriminative Classifier Methods:} we provide more analysis and justification on why and how the \stic learns class boundaries better than the baseline methods in Sec \ref{suppl_inn_vs_stic}.
    \item \textbf{More Qualitative Results:} We provide more qualitative results in Sec \ref{suppl_image_generation}.
\end{itemize}
%
\section{Additional Architecture Details and Algorithm}
\label{suppl_architecture_details}
\begin{figure}[!ht]
  \centering
  \includegraphics[height=0.3\textwidth,width=0.45\textwidth]{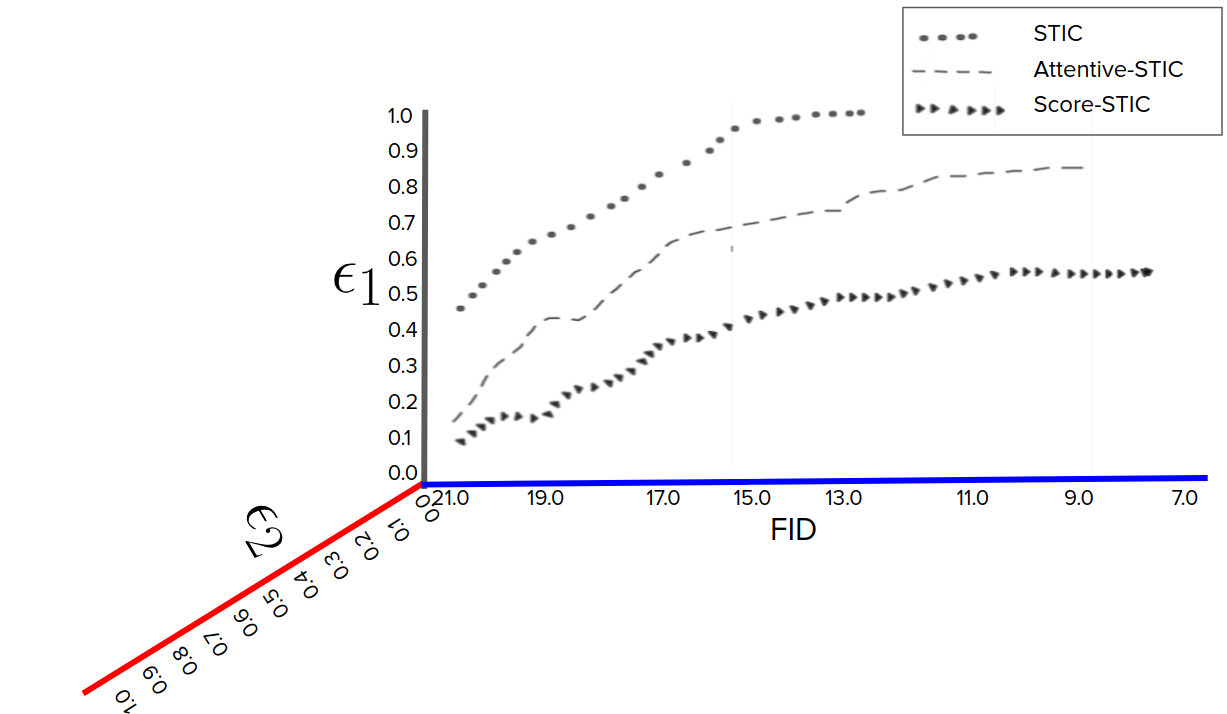}
  \caption{\textit{\textbf{$\epsilon_1$ vs. $\epsilon_2$ vs. FID:}} we observe the highest FID score (blue axis) if the scaling factors $\epsilon_1$ (black axis) and $\epsilon_2$ (red axis) are set to the range $\{0.9 - 1.0\}$ while keeping the range of other scaling factor in the range $\{0.01 - 0.02\}$. }
  \label{suppl_fig_1_FID_ep1_ep2}
  \vspace{-0.5cm}
\end{figure}

In addition to the methodology described in Sec \ref{sec_metodology}, we present the overall methodology in the form of an algorithm in \ref{alg:main}.
\paragraph{Further Discussion on \stic:} In continuation to the discussion in Sec \ref{sec_experiments_and_results}, we now provide more details of the experiment for \stic. The initial learning rate is set to 0.0001 while we have a learning rate decay of 0.3 after every $10k$ epochs. Varying $\epsilon_{1}$ between $\{0.9-1.0\}$ and $\epsilon_{2}$ between $\{0.9-1.0\}$ to the range produces good quality images with high FID (see Fig \ref{suppl_fig_1_FID_ep1_ep2}) potentially because the learning from stochastic gradient langevin dynamics at passes $\tau \in \{1, 2, \cdots, T\}$ and learning from Gram matrix similarity of real images to the synthetic images support each other's learning. However, varying $\epsilon_{3}$ between $\{0.01-0.02\}$ (i.e. lower range) helps to search the manifold and controls the diversity of synthesis. But, varying $\epsilon_{3}$ between $\{0.1-0.8\}$ (i.e. higher range) does not provide a good update signal. 

We choose WideResNet-28-10 due to various reasons, such as (1) we can directly drop the batch normalization without compromising the accuracy of the classifier; and (2) the architecture has less number of parameters, easy to train and widely used that has all the facilities of a standard ResNet architecture.

\begin{algorithm*}
\SetAlgoLined
\textbf{Input:} Number of passes $\tau\in\{1, 2, \cdots, T\}$, Minibatch size: $m$\\
\textbf{Output:} Trained \stic model\\
\For{a pass $\tau$}{
    iteration = 0\\
    \For{iteration $\leq 5k$}{
        Draw a batch of $K$ synthesized image-label pairs from previous classifier $p_{\theta^{\tau-1}}$ using GRMALA in Eqn \ref{eq_grmala}\;
        Draw a batch of $K$ synthesized \textit{mixup} image-label pairs from previous classifier $p_{\theta^{\tau-1}}$ using GRMALA in Eqn \ref{eq_grmala}\;
        Draw a batch of $N$ image-label pairs from dataset $p_{data}$\;
        Draw a batch of $N$ \textit{mixup} virtual image-label pairs similar to \cite{zhang2018mixup}\;
        Update weight $\theta^{\tau}$ of \stic classifier using mini-batch stochastic gradient descent with gradients as computed below:
        \[- \sum_{(x_i, y_i)\sim p_{data}}^{i=1,\cdots,N} \log p_{\theta^{\tau+1}}(y_i = y_{c}|x_i) - \sum_{(x^{mixup}_{k}, y^{mixup}_{k})\sim p_{mixup}}^{k=1,\cdots,K} \log p_{\theta^{\tau+1}}(y_{k}=y_{mixup}|x^{mixup}_{k}) \]
        \[- \sum_{(x_i, y_i)\sim p_{\theta^\tau}}^{i=1,\cdots,N} \log p_{\theta^{\tau+1}}(y_i = -1 |x_i) - \sum_{(x^{mixup}_{k}, y^{mixup}_{k})\sim p_{\theta^\tau}}^{k=1,\cdots,K} \log p_{\theta^{\tau+1}}(y_{k}= -1|x^{mixup}_{k})\]
    }
}
\caption{Training \stic}
\label{alg:main}
\end{algorithm*}
\paragraph{Further Discussion on Attentive\ stic:} Read $\mathcal{R}(\cdot)$ and LSTM network of \stic are followed from \cite{gregor2015draw}. 
The LSTM network provides a 100 dimensional feature vector to the discriminative classifier. 
The decoder network is the DCGAN, i.e. f100-conv1(8x8x1024)-conv2(64x64x512)-conv3(212x212x128)-conv4(512x512x3). 
The discrimnative classifier has the same hyperparamters as \stic.
%
\section{\stic vs. SOTA Image Synthesis using Discriminative Classifier Methods:}
\label{suppl_inn_vs_stic}
Our broad idea of using a discriminative classifier to synthesize class conditioned images may find similarity with earlier efforts \cite{jin2017introspective, lazarow2017introspective, grathwohl2019your}, however, in many aspects, they are different from our \stic methodology:
\begin{itemize}
    \item \textbf{\stic \textit{vs.} INN:} The INN methodology \cite{jin2017introspective} has $c\in\{1, 2, \cdots, C\}$ number of distinct CNN classifiers trained with ERM \cite{vapnik} and cascaded in a sequential manner. For any classifier, let's assume the $c^{th}$ classifier, the parameters are $W_{c} = \{\textbf{w}_{c}^{0}, \textbf{w}_{c}^{1_{1}}, \cdots, \textbf{w}_{c}^{1_{K}}\}$. The $\textbf{w}_{c}^{1}$ denotes the weights of the top $K$ separate layers for $K$ classes, while $\textbf{w}_{c}^{0}$ carries all internal features. The negative samples are sampled for each class. Such negative samples along with the real samples are then utilized by the c$+1^{th}$ classifier to segregate real samples to negative samples. On the other hand, the \stic classifier serves dual objectives, \textit{viz.} the interpolated samples from one class to another must be smooth and the classifier must learn tighter class boundaries so as to generate photo-realistic samples. Thus, \stic is different from INN in several ways, such as: (1) \stic is trained with VRM \cite{zhang2018mixup} with virtual image-label pairs along with real image-label pairs that provide a good learning of smooth class boundaries and tighter class boundary across passes; (2) the loss function and the training methodology of \stic is different from INN, i.e. INN uses a separate branch for each classes, but, \stic instead uses a single architecture wide ResNet (ref. Sec \ref{sec_experiments_and_results}) and trains the method; (3) \stic optimizes less number of parameters (single architecture) than INN (classifier with multiple branches), and, we note that, the convergence time and image quality of \stic is far better than INN due to training the  classifier with VRM; (4) utilization of synthetic samples as fake sample is different from INN; and most importantly (5) our sampling technique, i.e. GRMALA (ref Sec \ref{sec_metodology}), is novel and different from the MCMC-based sampling of INN.
    \item \textbf{JEM vs \stic:} The JEM \cite{grathwohl2019your} methodology is developed based on an energy based estimation of $p(x)$ and $p(x,y)$. We note that such a method is different from ours, as: (1) as described above, the \stic uses VRM based training and GRMALA. (2) The recurrent class boundary re-estimation way of training is different from the JEM methodology.
\end{itemize}

\noindent We ask ourselves the question, \textit{how does VRM help the \stic method?} From the understanding of VC theory \cite{vapnik}, the classification error of a classifier $\hat{f}$ can be decomposed as:
\begin{equation}
\label{suppl_eq_vapnik}
    R(\hat{f}) - R(f) \leq O\Big{(}\frac{|\hat{\mathcal{F}}|_{C}}{n^{\alpha}}\Big{)} + \epsilon
\end{equation}
here, $f\in\mathcal{F}$ is the true classifier function we wish to approximate using the function $\hat{f}\in \hat{\mathcal{F}}$. The $|\cdot|_{C}$ is the class capacity measure, error is the $R$, number of data points are shown as $n$ and $\alpha$ is the learning rate. We note that, the $\epsilon$ is the approximation error of $\hat{\mathcal{F}}$ with respect to the function $\mathcal{F}$. To this end, a loss function $l(\cdot)$ penalizes the difference between the predictions $\hat{f}(x)$ and the ground truth $y$ sampled from $p_{data}(x,y)$. The average of the loss function $l(\cdot)$ is averaged over training data samples and the empirical risk is minimized as follows:
\begin{equation}
\label{suppl_eq_erm}
    \begin{split}
    & \qquad R(\hat{f}) =  \sum_{x_i, y_i \in p_{data}(x,y)}^{i=1,\cdots,n} l(\hat{f}(x_{i}), y_{i})
    \end{split}
\end{equation}
%
\begin{figure*}[!ht]
    \centering
    \includegraphics[width=\textwidth]{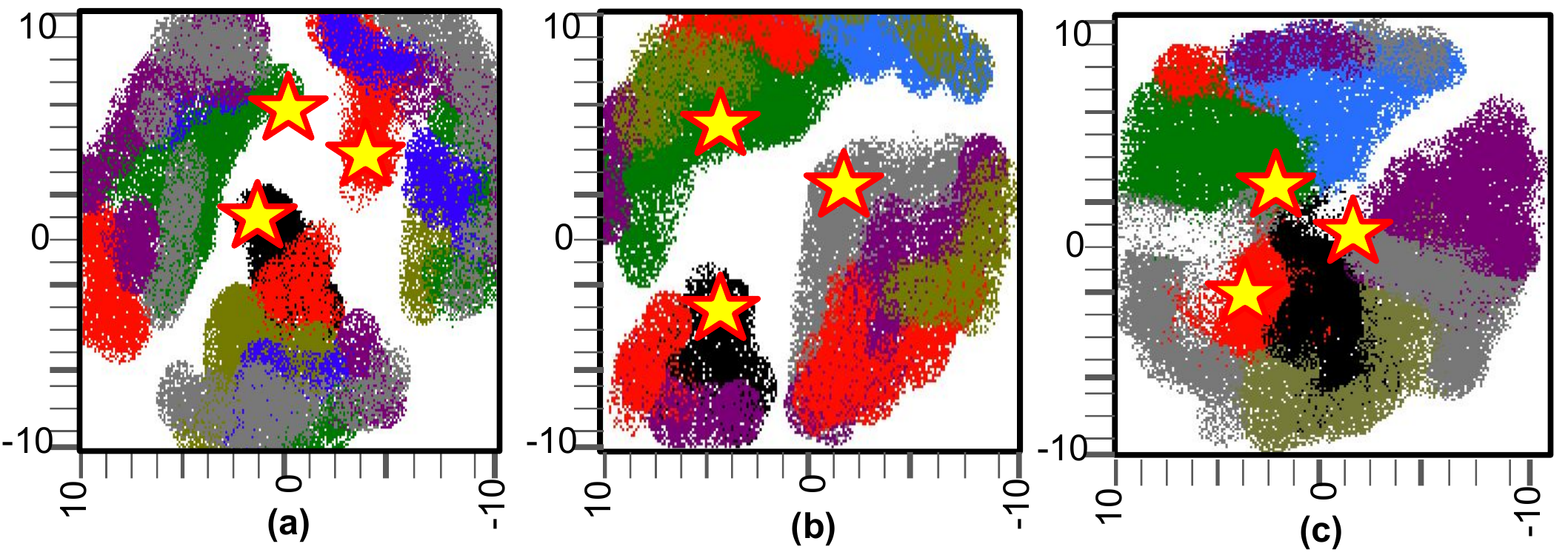}
    \caption{\textit{\textbf{Visualization of CIFAR 10 dataset Class Boundaries}}: We visualize seven class (i.e. airplane, automobile, bird, cat, deer, dog, frog classes of CIFAR 10 are shown to reduce the clutter) boundaries of (a) INN (b) JEM and (c) \stic on the CIFAR 10 dataset. We observe that the class boundary is very compact in \stic. The yellow stars are points we sample and synthesize images.}
    \label{suppl_fig_visualization}
\end{figure*}
%
A classifier function $\hat{f}$ trained with \stic takes the following form:
\begin{equation}
\label{suppl_eq_vrm}
\begin{split}
    & \quad R(\hat{f}) =  \sum_{(x_i, y_i)\sim p_{data}}^{i=1,\cdots,N} l(\hat{f}(x_{i}), y_{i}) \\
    & \quad + \sum_{(x^{mixup}_{k}, y^{mixup}_{k})\sim p_{mixup}}^{k=1,\cdots,K} l(\hat{f}(x_{k}^{mixup}), y_{k}^{mixup}) \\
    & \quad + \sum_{(x_i, y_i)\sim p_{\theta^\tau}}^{i=1,\cdots,N} l(\hat{f}(x_{i}), -1)\\
    & \quad + \sum_{(x^{mixup}_{k}, y^{mixup}_{k})\sim p_{\theta^\tau}}^{k=1,\cdots,K} l(\hat{f}(x_{k}^{mixup}), -1)
\end{split}
\end{equation}
\noindent Similar to the argument presented in \cite{chapelle2000vicinal}, if the virtual image-labels are a poor approximation of class vicinity then \stic trained with VRM performs at least as good as a classifier trained with ERM. We note that the virtual image-labels using mixup of softmax \cite{zhang2018mixup} provides a good approximation of class vicinity. In addition to that, the recurrent self-estimation with VRM is a better approximation of class vicinity w.r.t the method proposed in \cite{zhang2018mixup}. We show the class boundary visualization of INN, JEM and \stic in Fig \ref{suppl_fig_visualization} and we note that the \stic class boundary is compact - supporting our claim. We also note that, the use of GRMALA based synthesis also provides good learning signal to estimate class boundaries.  
%
\section{More Synthesized Images}
\label{suppl_image_generation}
\begin{figure*}[!ht]
    \centering
    \includegraphics[width=\textwidth,height=0.7\textwidth]{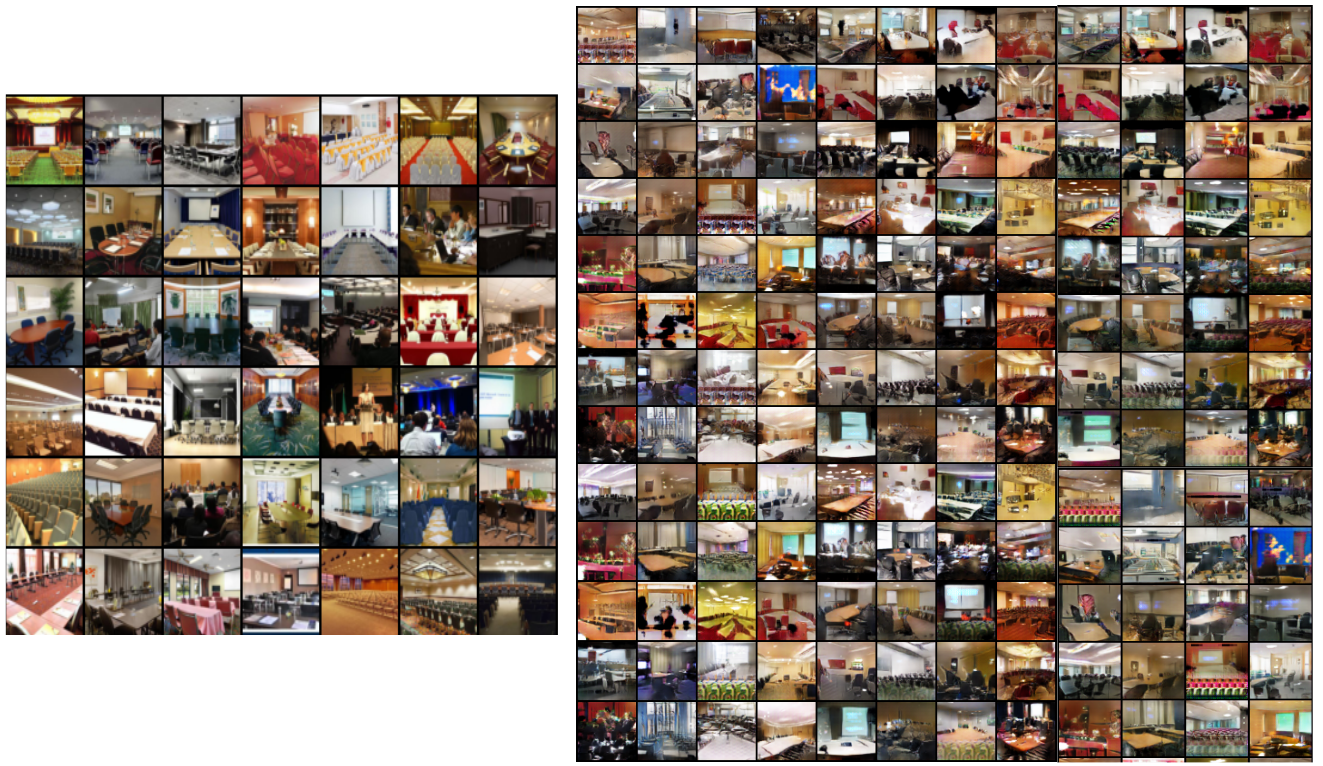}
    \captionof{figure}{\textit{\textbf{More Qualitative Results on the LSUN dataset:}} We show qualitative results on the LSUN conference class.}
     \label{suppl_fig_lsun_conferecne}
\end{figure*}

\begin{figure*}[!ht]
    \centering
    \includegraphics[width=\textwidth,height=0.7\textwidth]{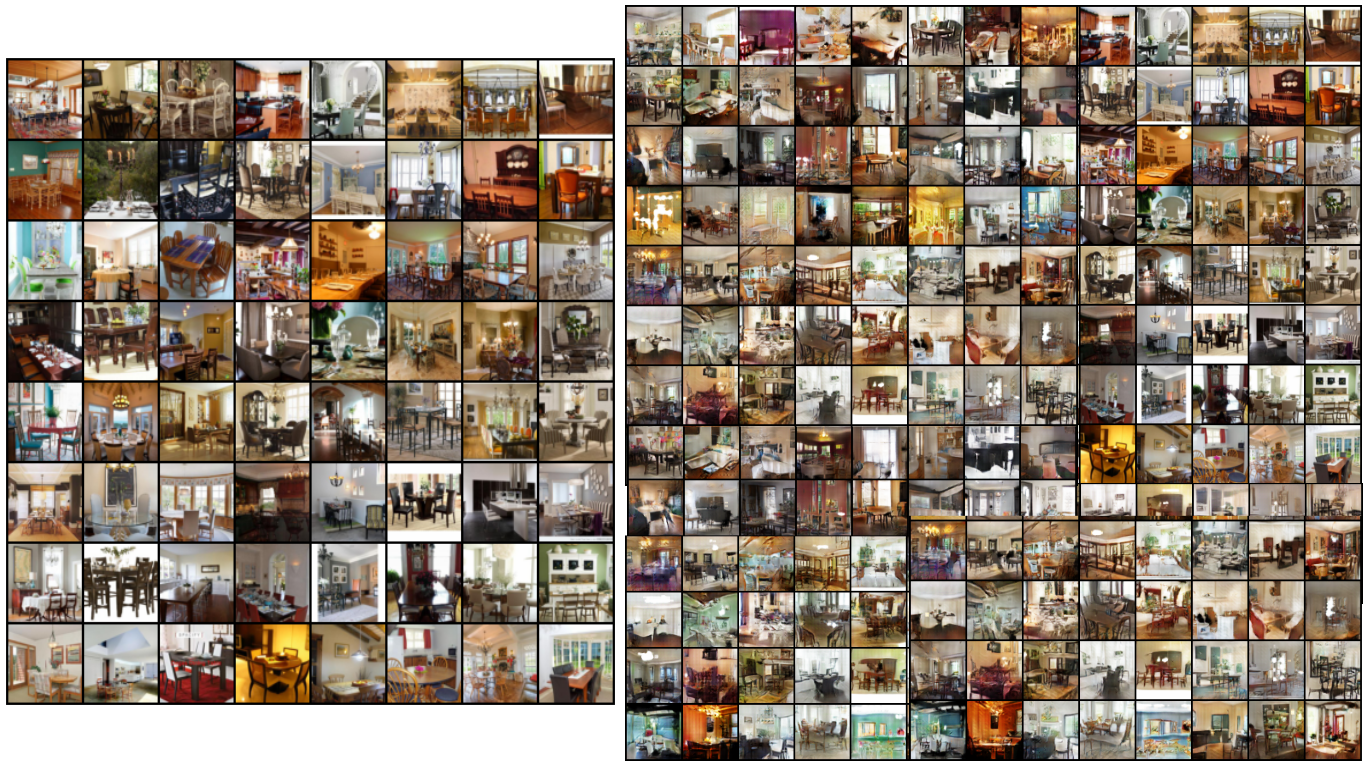}
    \captionof{figure}{\textit{\textbf{More Qualitative Results on LSUN dataset:}} We show qualitative results on LSUN dinning hall class.}
     \label{suppl_fig_lsun_dining_room}
\end{figure*}

\begin{figure*}[!ht]
    \centering
    \includegraphics[width=\textwidth,height=0.7\textwidth]{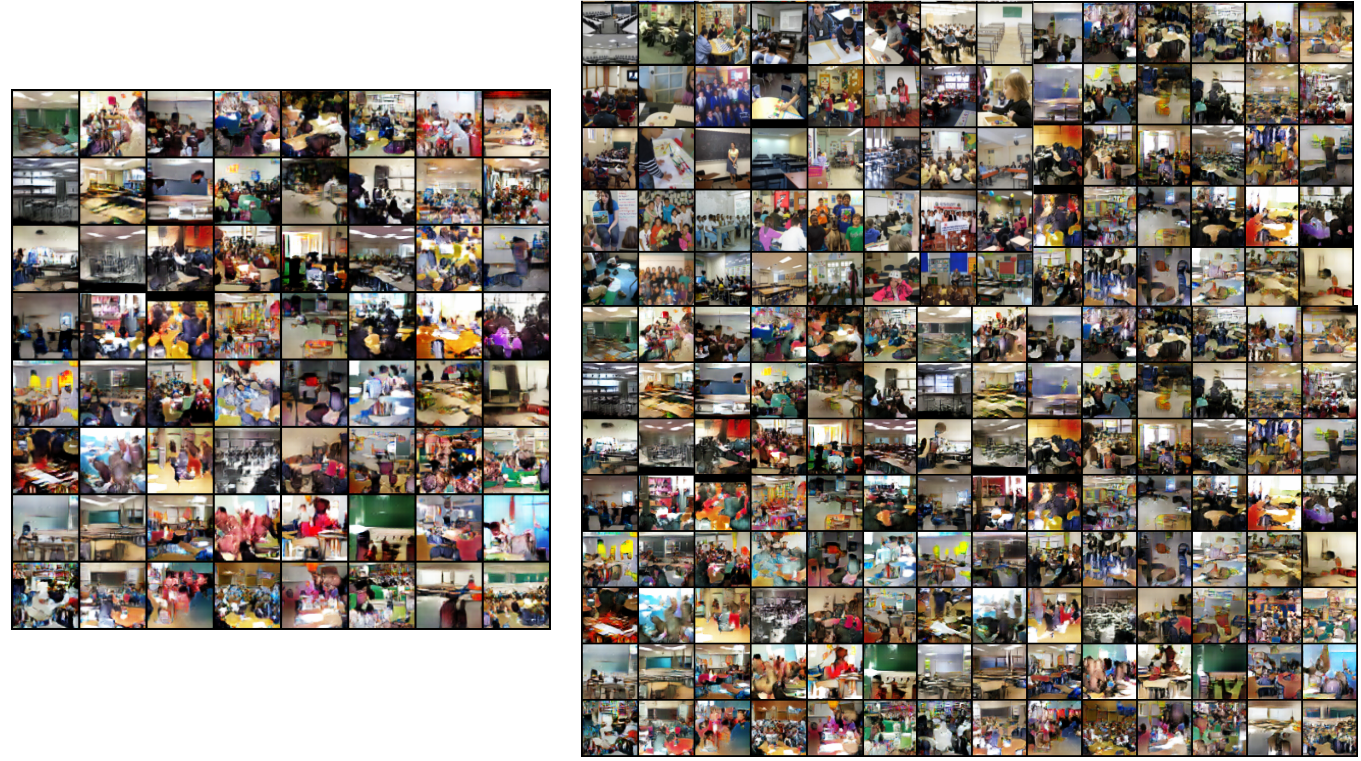}
    \captionof{figure}{\textit{\textbf{More Qualitative Results on LSUN dataset:}} We show qualitative results on LSUN classroom class.}
     \label{suppl_fig_lsun_classroom}
\end{figure*}

\begin{figure*}[!ht]
    \centering
    \includegraphics[width=0.8\textwidth,height=1.12\textwidth]{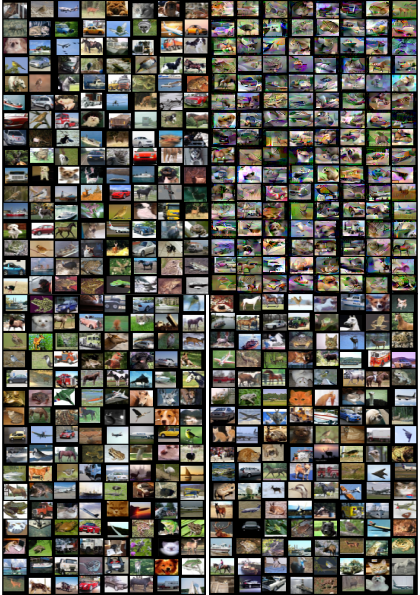}
    \captionof{figure}{\textit{\textbf{More Qualitative Results on CIFAR 10 Dataset:}} We show qualitative results on CIFAR 10 images (mixed classes).}
     \label{suppl_fig_cifar_10}
\end{figure*}

\begin{figure*}[!ht]
    \centering
    \includegraphics[width=\textwidth,height=\textwidth]{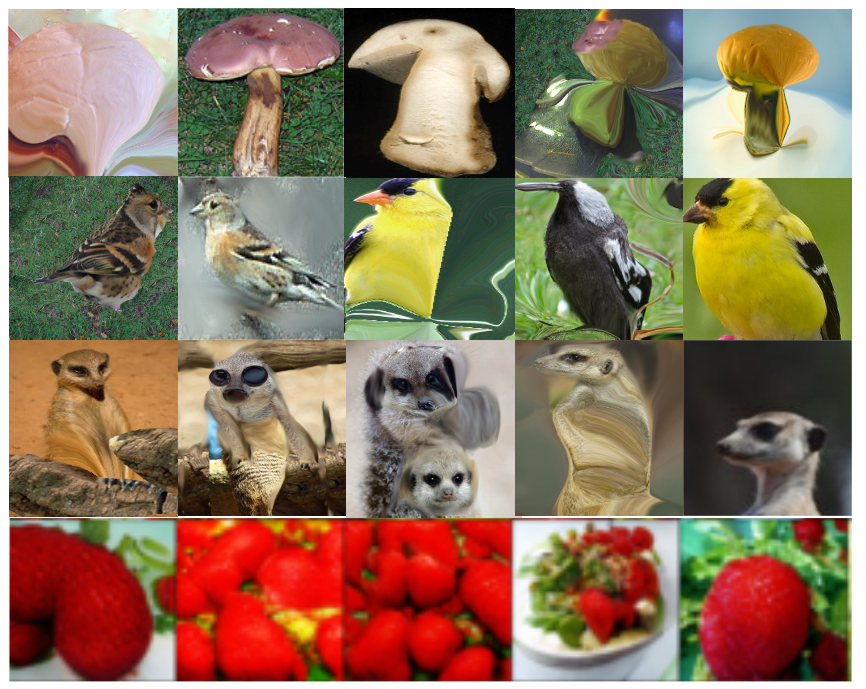}
    \captionof{figure}{\textit{\textbf{More Qualitative Results on ImageNet Dataset:}} We show qualitative results on ImageNet images.}
     \label{suppl_fig_imagenet}
\end{figure*}

In addition to our qualitative results shown in Fig \ref{fig_2_intro_images}, in this section we show more qualitative images of LSUN, Cifar 10 and ImageNet datasets in Figs \ref{suppl_fig_lsun_conferecne}-\ref{suppl_fig_imagenet} (please see next pages).
%
\section{Synthesizing using \stic}
\label{suppl_synthesize_using_stic}
\begin{figure*}[!ht]
    \centering
    \includegraphics[width=\textwidth,height=0.6\textwidth]{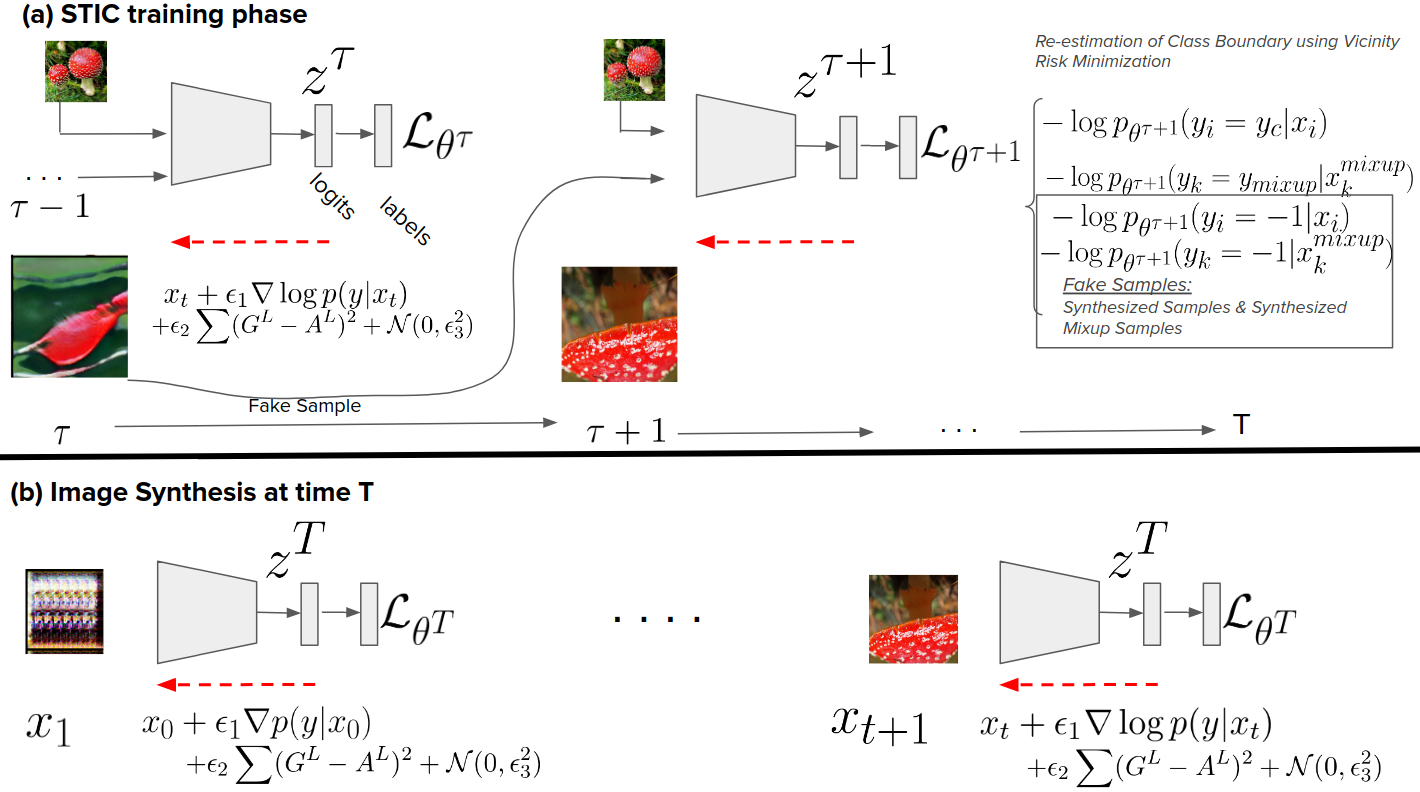}
    \captionof{figure}{\textit{\textbf{Image Synthesis using STIC at Image Generation Phase:}} (a) training phase of \stic, (b) image synthesis from \stic at time $t$.}
     \label{suppl_fig_stic_training}
\end{figure*}

Following the training process described in Sec \ref{sec_metodology} and Algorithm \ref{alg:main}, \stic synthesize images as follows: starting with an initial $x_{0}$ typically sampled from a Gaussian distribution $\mathcal{N}(0,I)$, the GRMALA uses the transition operator, \textit{viz.} $x_{t+1} = x_{t} + \epsilon_{1}\nabla\log p(x_{t}) + \sum (G^{L}(x_{t}) - A^{L}(x_{t}))^{2} +\mathcal{N}(0,\epsilon_{2}^{2})$, synthesize novel image samples from the classifier at pass $\tau = T$, see Fig \ref{suppl_fig_stic_training} (b). We show the training process again in Fig \ref{suppl_fig_stic_training} (a).

%
%
%
%
%
%
%
\begin{figure*}[!ht]
    \centering
    \includegraphics[width=\textwidth,height=0.5\textwidth]{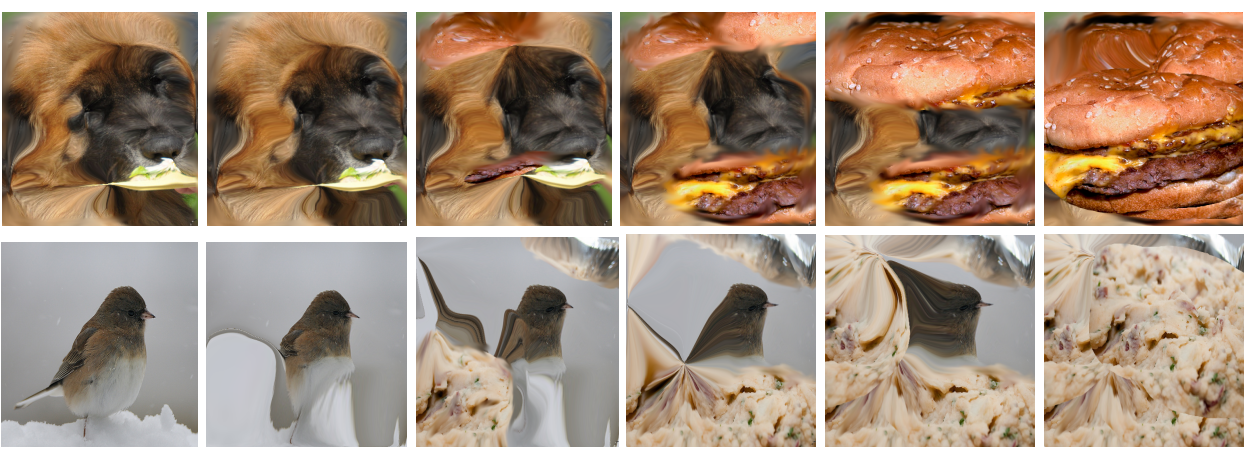}
    \captionof{figure}{\textit{\textbf{Class Interpolation Results on ImageNet Dataset:}} We show two more interpolation results on ImageNet images.}
     \label{suppl_fig_interpolation}
\end{figure*}
%